\documentclass[]{baidu}

\usepackage[toc,page,header]{appendix}

\usepackage{natbib}

\usepackage{multicol} 
\usepackage{enumitem} 

\usepackage{tcolorbox}
\usepackage{xcolor}
\usepackage{listings} 
\tcbuselibrary{skins, breakable}

\newtcolorbox{modelresponse}[1][]{
  colback=blue!5!white,
  colframe=blue!75!black,
  title=\textbf{Model Response (Translated)},
  fonttitle=\bfseries,
  breakable,
  #1
}

\newtcolorbox{modelresponse_raw}[1][]{
  colback=blue!5!white,
  colframe=blue!75!black,
  title=\textbf{Model Response},
  fonttitle=\bfseries,
  breakable,
  #1
}

\newtcolorbox{modelthink}{
  colback=gray!10!white,
  colframe=gray!50!black,
  title=\textit{Internal Reasoning (Translated)},
  fonttitle=\small,
  fontupper=\small\color{gray!30!black},
  breakable,
  boxrule=0.5pt
}

\usepackage{enumitem}
\usepackage{xcolor}
\usepackage{listings}

\lstdefinestyle{jsonstyle}{
    basicstyle=\ttfamily\footnotesize, 
    breaklines=true,     
    frame=single,        
    numbers=left,        
    numberstyle=\tiny,   
    backgroundcolor=\color{gray!5}, 
    keywordstyle=\color{blue},      
    stringstyle=\color{red}         
}

\usepackage{CJKutf8}

\usepackage{xargs}  

\usepackage{todonotes}  

\usepackage{multirow}

\usepackage{cleveref}

\usepackage{amsmath}
\usepackage{dsfont}
\usepackage{amssymb} 

\usepackage{subcaption}
\usepackage{svg}

\usepackage{fontawesome}
\usepackage[utf8]{inputenc}
\usepackage{booktabs}
\usepackage{graphicx}
\usepackage{array}
\usepackage{tabularx}
\usepackage{xspace}
\usepackage{threeparttable} 

\usepackage{makecell}
\usepackage{wrapfig}
\usepackage{amsmath}
\usepackage{diagbox}
\usepackage{booktabs}
\usepackage{float} 
\usepackage{colortbl}    
\usepackage{xspace}

\usepackage{soulutf8} 
\definecolor{nred}{RGB}{196, 38, 11}
\definecolor{ngreen}{RGB}{18, 141, 21}
\definecolor{nblue}{RGB}{41, 52, 190}
\definecolor{hzw}{RGB}{223, 97, 76}
\definecolor{lt}{RGB}{54, 89, 170}

\definecolor{zlblue}{RGB}{196, 223, 251}

\usepackage{tikz}       

\tcbuselibrary{skins} 
\tcbset{
  aibox/.style={
    width=\linewidth,
    top=8pt,
    bottom=4pt,
    colback=blue!6!white,
    colframe=black,
    colbacktitle=black,
    center,
    enhanced, 
    attach boxed title to top left={yshift=-0.1in,xshift=0.15in},
    boxed title style={boxrule=0pt,colframe=white,},
  }
}
\newtcolorbox{AIbox}[2][]{aibox,title=#2,#1}

\usepackage{minitoc}

\newtcolorbox{promptbox}[2][Prompt]{
colback=black!5!white,
arc=5pt, 
boxrule=0.5pt,
fonttitle=\bfseries,
title=#1, 
before upper={\small}, fontupper=\fontfamily{ptm}\selectfont,
colframe=#2, 
}
\definecolor{ogreen}{RGB}{34, 139, 34}

\title{
QianfanHuijin Technical Report
\vspace{0.5em}

    \Large A Novel Multi-Stage Training Paradigm for Finance Industrial LLMs 
}

\author[]{Qianfan Team, Baidu AI Cloud}

\contribution[\dagger]{Corresponding authors}

\abstract{
Domain-specific enhancement of Large Language Models (LLMs) within the financial context has long been a focal point of industrial application. While previous models such as BloombergGPT and Baichuan-Finance primarily focused on knowledge enhancement, the deepening complexity of financial services has driven a growing demand for models that possess not only domain knowledge but also robust financial reasoning and agentic capabilities. In this paper, we present QianfanHuijin, a financial domain LLM, and propose a generalizable multi-stage training paradigm for industrial model enhancement.

Our approach begins with Continual Pre-training (CPT) on financial corpora to consolidate the knowledge base. This is followed by a fine-grained Post-training pipeline designed with increasing specificity: starting with Financial SFT, progressing to Finance Reasoning RL and Finance Agentic RL, and culminating in General RL aligned with real-world business scenarios. Empirical results demonstrate that QianfanHuijin achieves superior performance across various authoritative financial benchmarks. Furthermore, ablation studies confirm that the targeted Reasoning RL and Agentic RL stages yield significant gains in their respective capabilities. These findings validate our motivation and suggest that this fine-grained, progressive post-training methodology is poised to become a mainstream paradigm 
for various industrial-enhanced LLMs.
}

\date{January 1, 2026}

\checkdata[Availability]{The QianfanHuijin/QianfanHuijin-Reason model series are hosted on the Qianfan ModelBuilder Platform.}

\begin{document}
\maketitle

\newpage

\section{introduction}




Large Language Models (LLMs) have demonstrated immense potential across diverse industries. Pioneering works like BloombergGPT ~\citep{wu2023bloomberggpt} and Baichuan-Finance ~\citep{zhang2024baichuan4} have demonstrated the value of domain-specific pre-training. 
Nevertheless, the financial scenarios demand strict reliability, deep domain expertise, and regulatory compliance. Due to its data-intensive nature and the requirement for rigorous logic, general-purpose LLMs often face significant limitations:

\begin{enumerate}
    \item \textbf{Domain Knowledge Gap:} They lack a deep understanding of specialized financial terminology, complex concepts, and inherent market logic.
    \item \textbf{Reasoning \& Computational Precision:} When handling tasks like financial statement analysis or numerical calculation, generic models are prone to factual errors and hallucinations.
    \item \textbf{Compliance \& Security Risks:} Ensuring generated content strictly adheres to rigorous financial regulations remains a significant challenge.
    \item \textbf{Tool Dependency:} Real-world financial tasks often require interoperability with external databases and calculation engines, a capability often missing in general models.
\end{enumerate}

To address these challenges, we present \textbf{QianfanHuijin}, a series of large language models tailored for the financial domain. This report details the full technology stack of QianfanHuijin, developed to construct a specialized model with deep domain knowledge, superior reasoning capabilities, robust tool utilization, and high compliance. Our core contribution lies in designing and implementing an end-to-end optimization paradigm, spanning from data engineering to model post-training.

\begin{figure}[!h]
    \centering
    \includegraphics[width=0.9\linewidth]{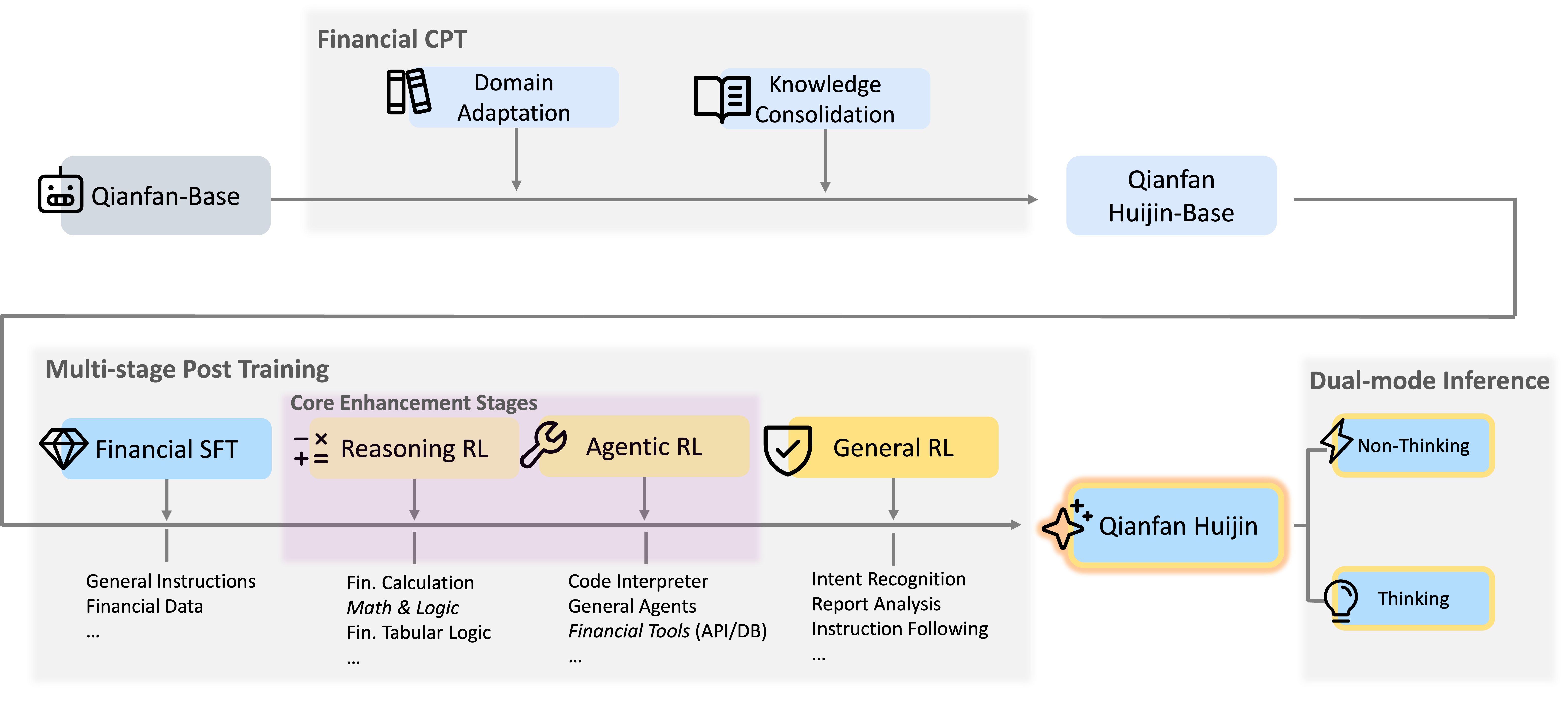}
    \caption{Illustration of the multi-stage training framework of QianfanHuijin}
    \label{fig:overview}
\end{figure}

The main contributions of this work are as follows:

\begin{itemize}
    \item \textbf{Phased Continual Pre-training (CPT):} We adopt a curriculum learning paradigm consisting of ``Financial Knowledge Injection'' and ``Financial Capability Enhancement''. This strategy enables the model to absorb financial knowledge in a coarse-to-fine manner while retaining its general language capabilities and preliminarily aligning with instruction intent.

    \item \textbf{Multi-stage Post-training Pipeline:} Recognizing that the traditional \textit{SFT $\to$ General RL} paradigm falls short in meeting the reasoning demands of complex financial tasks, we designed a progressive workflow: \textbf{SFT $\to$ Reasoning RL $\to$ Agentic RL $\to$ General RL}, as illustrated in Figure \ref{fig:overview}.
    \textit{Reasoning RL} focuses on cultivating logical rigor and computational precision, while \textit{Agentic RL} reinforces tool collaboration. These intermediate stages lay a solid foundation for the final \textit{General RL} stage, significantly raising the ceiling of the model's professional capabilities. Furthermore, we integrated a dual-mode mechanism—\textbf{``Thinking'' and ``Non-thinking''}—allowing the model to flexibly switch between high-efficiency responses for simple tasks and explicit reasoning for complex analysis.

    \item \textbf{Controllable Instruction Synthesis Framework for Finance (CIS-F):} To overcome bottlenecks in the diversity, complexity, and accuracy of financial instruction data, we propose CIS-F. This system utilizes knowledge-driven generation, multi-layer verification, and instruction evolution to systematically enhance data quality and reliability.

    \item \textbf{Dual-verifier Reward Model:} We constructed a reward system combining a \textbf{Rule-based Verifier} and an \textbf{LLM-based Verifier}. This provides precise, interpretable reward signals for different task types, significantly improving reinforcement learning efficiency and model reliability.
\end{itemize}

Extensive empirical results demonstrate that QianfanHuijin significantly outperforms leading models of similar size on authoritative benchmarks.

Finally, to address the diverse demands of real-world financial applications—ranging from latency-sensitive retrieval to accuracy-critical analysis—we integrate a flexible dual-mode mechanism ("Thinking" vs. "Non-thinking"). This design allows Huijin to dynamically trade off between rapid response for routine queries and deep, explicit reasoning for complex problems, maximizing its practical value in industrial deployment.
\newpage

\section{Continual Pre-training}
\subsection{Data Pipelines}

\subsubsection{Data Cleaning and Filtering}
The raw financial data is primarily sourced from two channels:
\begin{itemize}
    \item \textbf{Web Data:} Financial-related content filtered from general common crawl data.
    \item \textbf{Proprietary Financial Data:} Exclusive domain-specific data including institutional research, real-time market feeds, and detailed financial statements obtained via strategic partnerships. These datasets represent private-domain resources not available in the public domain.
\end{itemize}

\begin{figure}[h]
    \centering
    \includegraphics[width=0.9\linewidth]{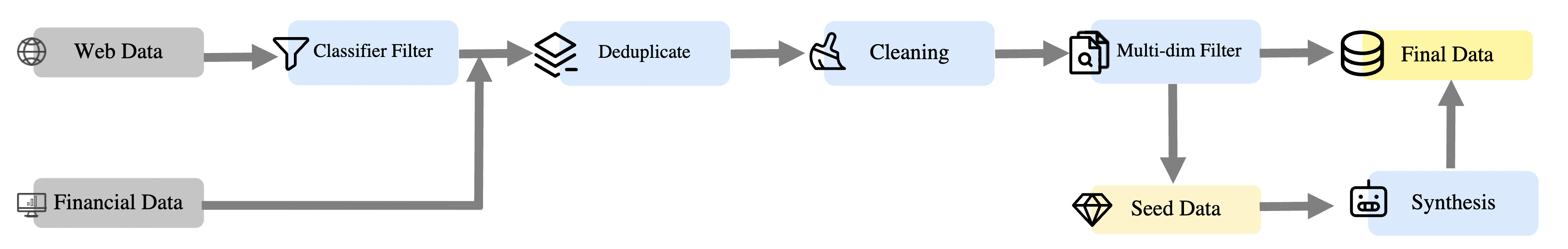}
    \caption{The Data Cleaning and Filtering Pipeline}
    \label{fig:data_pipeline}
\end{figure}

To construct a high-quality corpus, we implemented a multi-stage filtering pipeline (see Figure~\ref{fig:data_pipeline}). First, we trained a lightweight \textbf{BERT-based classifier} to filter financial-relevant content from the general pool. Subsequently, we applied rule-based cleaning operators to remove HTML tags, advertisements, and overly short text segments.

Furthermore, inspired by methods such as CCI-4.0~\cite{liu2025cci4}, we employed a \textbf{multi-dimensional quality scoring system}. This system evaluates data based on textual quality and knowledge density. Only data exceeding a strict threshold is retained. From this high-quality pool, we further sampled a subset to serve as seed data for synthesis. The final training corpus consists of approximately tens of billions of tokens high-density, high-quality financial text.


\subsubsection{Data Synthesis}





\begin{figure}[H] 
    \centering
    \begin{minipage}[c]{0.55\textwidth} 
    \vspace{-3mm}
    To further enhance the model's ability to understand and apply knowledge, we developed a synthesis pipeline combining \textbf{Self-QA} ~\cite{zhang2023self} and \textbf{Evol-Instruct}~\cite{luo2023wizardmath} methods. Figure \ref{fig:data_synthesis} illustrates the detailed steps of our data synthesis workflow. The process is as follows:
        \begin{enumerate}
            \item \textbf{Knowledge Extraction \& Seeding:} We first extract key financial knowledge points from the high-density corpus. We then generate questions by integrating typically 3-5 randomly selected knowledge points to ensure the questions cover multiple aspects of the context.
            \item \textbf{Instruction Evolution:} To increase difficulty and breadth, we apply multi-turn instruction evolution to refine the questions.
            \item \textbf{Consistency Verification:} We utilize advanced models (e.g., DeepSeek-R1 ~\citep{guo2025deepseek}, Ernie-4.5 ~\citep{ernie2025technicalreport}) to generate multiple responses for the same synthetic question. We then apply a \textbf{Semantic Consistency Filter} to remove QA pairs where the key information across responses is contradictory.
        \end{enumerate}
    \end{minipage}%
    \hfill
    \begin{minipage}[c]{0.39\textwidth}
        \centering
        \vspace{5mm}
        \includegraphics[width=\linewidth]{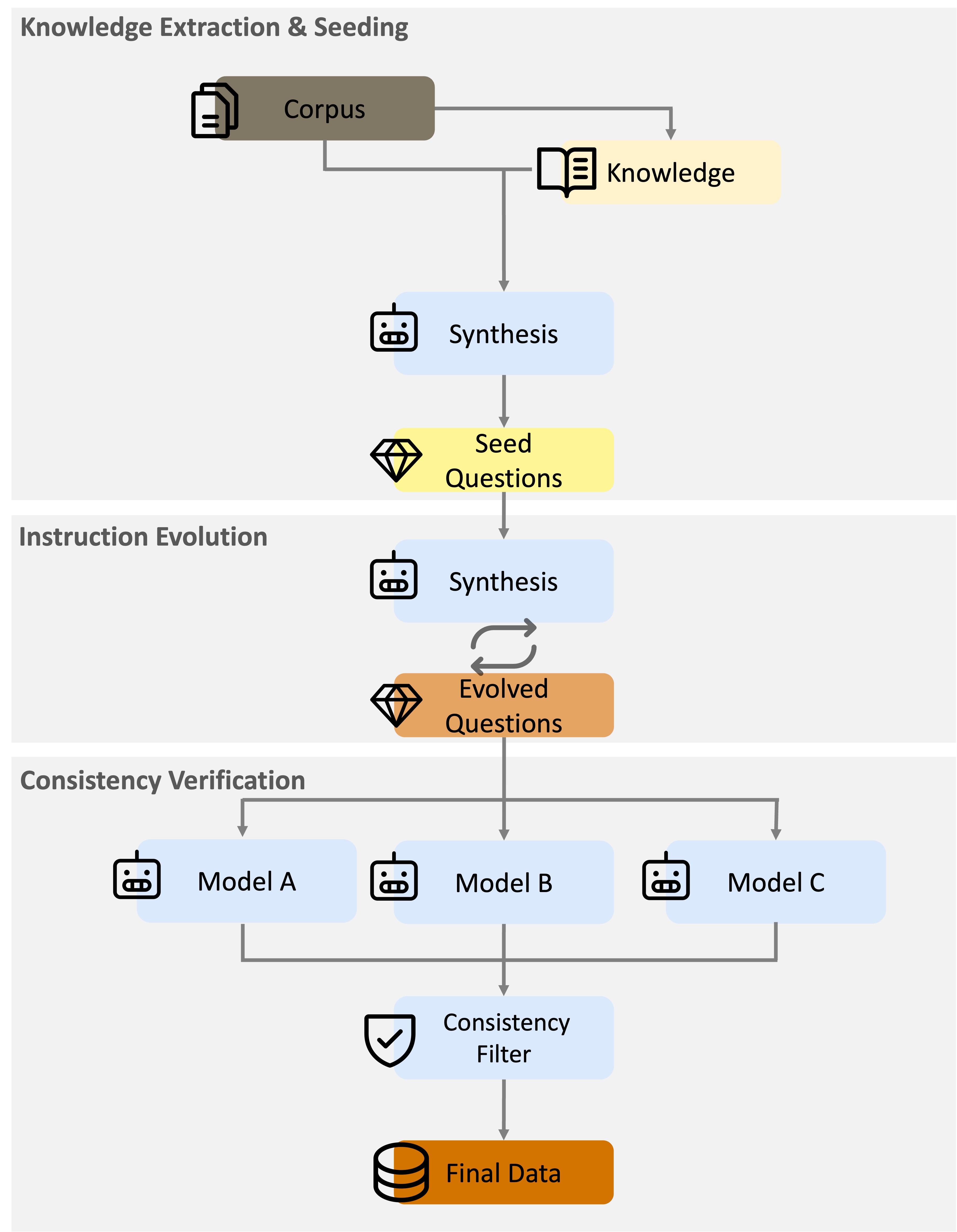}
        \caption{The Data Synthesis Workflow} 
        \label{fig:data_synthesis}
    \end{minipage}
\end{figure}

\subsection{Training Strategy}
To build a specialized LLM with deep semantic understanding and precise Q\&A capabilities in finance, we designed a \textbf{Two-Stage Continual Pre-training (CPT) Strategy}. This strategy follows the philosophy of ``Curriculum Learning,'' enabling the model to absorb knowledge from coarse to fine (see Table~\ref{tab:cpt_results}).

\begin{figure}[!h]
    \centering
    \includegraphics[width=0.8\linewidth]{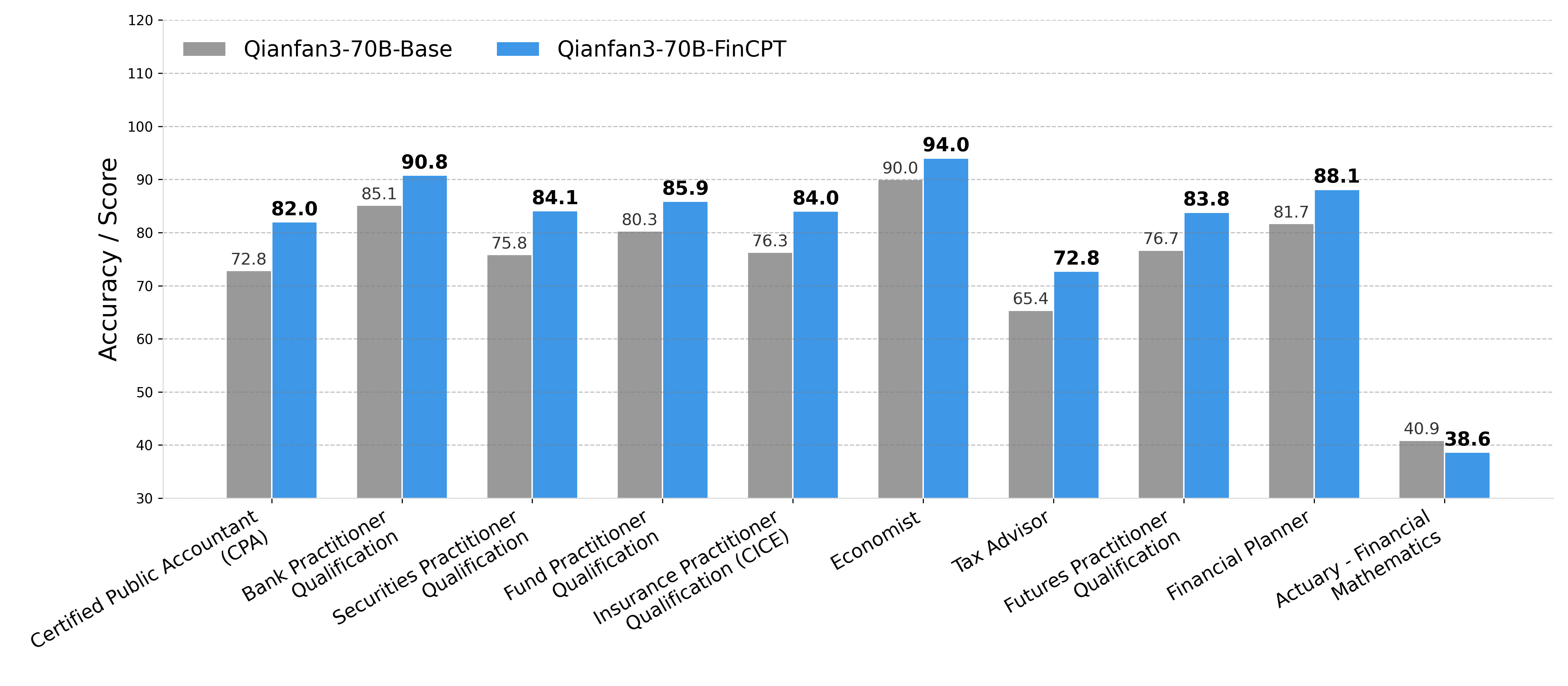}
    \caption{FinanceIQ Submetrics}
    \label{fig:FinanceIQ}
\end{figure}

\begin{itemize}
    \item \textbf{Stage 1: Financial Knowledge Injection (Foundation Building).} 
    The primary goal is to transform the general LLM into a financial expert foundation model. We focus on mastering extensive financial terminology, concepts, logical relations, and writing styles. In this stage, we train on a large-scale mixture of \textbf{Financial and General corpora}. This ensures effective domain adaptation while preventing catastrophic forgetting of general language capabilities.

    
    \item \textbf{Stage 2: Financial Capability Enhancement (Knowledge Consolidation).} 
    Building upon the Stage 1 model, this phase prioritizes \textbf{deepening knowledge} and optimizing the \textbf{distribution of specialized domain knowledge}. 
    We curate a higher-density dataset by sampling top-tier corpora from Stage 1 and up-weighting \textbf{Financial Q\&A data}.
    This strategy ensures a strong balance between specialized financial insights and general capabilities, effectively bridging the gap between pre-training and the subsequent Supervised Fine-Tuning (SFT) phase. The performance significantly improved after Stage 2(see Figure \ref{fig:FinanceIQ}).
\end{itemize}

\begin{table}[!h]
    \centering
    \caption{Performance Comparison Across Training Stages}
    \label{tab:cpt_results}
    \begin{tabular}{lccc}
        \toprule
        \textbf{Model} & \textbf{FinanceIQ} & \textbf{OpenFinData} & \textbf{Flamer-Cer} \\
        \midrule
        Qianfan3-70B-Base & 74.50 & 73.31 & 73.68 \\
        Qianfan3-70B-FinCPT-Stage1 & 79.59 & 86.88 & 78.00 \\
        \textbf{Qianfan3-70B-FinCPT-Stage2} & \textbf{80.43} & \textbf{88.45} & \textbf{80.00} \\
        \bottomrule
    \end{tabular}
\end{table}

\section{Post-Training}

\subsection{Overview}
Post-training for pre-trained models typically follows a two-stage paradigm: Supervised Fine-Tuning (SFT) followed by General Reinforcement Learning (General RL). While effective for general conversation and instruction following, we identified a critical limitation when applying this paradigm to the financial domain: \textbf{General RL fails to effectively bridge the significant reasoning gap between the SFT model and complex financial tasks.}

Financial tasks---such as financial statement analysis, ratio calculations, portfolio optimization, and valuation---demand not only factual accuracy but also rigorous logical chains, precise numerical computation, and collaboration with external tools. Directly applying General RL often leads to a performance plateau, as the model struggles to self-improve on these specialized capabilities.

To overcome this bottleneck, we redesigned the pipeline by inserting two targeted enhancement stages between SFT and General RL: \textbf{Reasoning RL (Stage 2)} and \textbf{Agentic RL (Stage 3)}.

\begin{itemize}
    \item \textbf{Reasoning RL:} Utilizes large-scale, high-quality data focused on financial logic and calculation. The goal is to deepen the ``imitative reasoning'' learned during SFT into a truly internalized logical and computational capability.
    \item \textbf{Agentic RL:} Builds upon the reasoning foundation to empower the model with the ability to invoke external databases and calculation engines, following the ReAct paradigm~\cite{yao2022react, schick2023toolformer}.
\end{itemize}

Furthermore, considering the diverse nature of financial applications---ranging from tasks favoring fast, straightforward responses to those involving cautious, in-depth decision analysis---we integrated a dual-mode mechanism in the SFT stage: \textbf{``Thinking'' and ``Non-thinking'' modes}. This allows the model to intelligently assess task complexity, providing direct answers for simple queries while explicitly unfolding reasoning steps for complex problems, thereby optimizing the trade-off between efficiency and quality.

\begin{figure}
    \centering
    \includegraphics[width=0.95\linewidth]{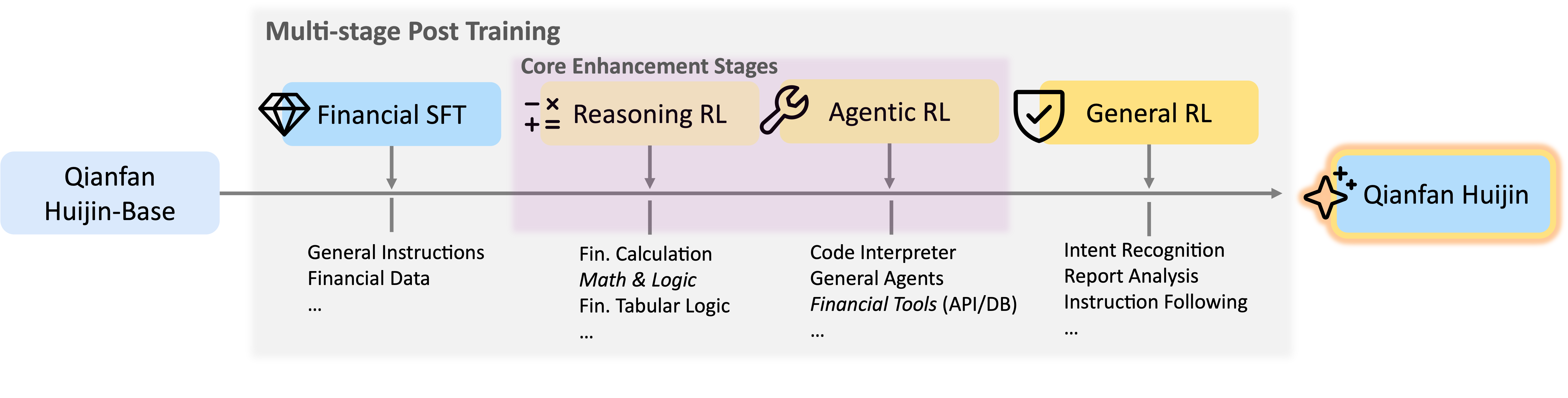}
    \caption{Illustration of 4 Stages}
    \label{fig:4stages}
\end{figure}

In summary, we propose a four-stage \textbf{Progressive Post-training Pipeline}:

\begin{enumerate}
    \item \textbf{Stage 1 (SFT):} Establish foundational language understanding and instruction adherence, incorporating both Thinking and Non-thinking modes.
    \item \textbf{Stage 2 (Reasoning RL):} Focus on finance and mathematics to cultivate rigor in deep reasoning and computation.
    \item \textbf{Stage 3 (Agentic RL):} Target real-world applications by reinforcing collaboration with external tools.
    \item \textbf{Stage 4 (General RL):} Comprehensive optimization to align specialized capabilities with general human preferences.
    The detailed workflow of this four-stage pipeline is illustrated in Figure \ref{fig:4stages}.
\end{enumerate}

This systematic flow transforms the knowledge potential of the pre-trained model into reliable capabilities for solving real-world financial problems.

\subsection{RL Data Synthesis Strategy}

High-quality, specialized data is the decisive factor for the capability ceiling in RL training. However, the financial domain presents unique dilemmas: real-world data is often ``declarative'' (e.g., reports, news) rather than ``instructional'' (Q\&A), and expert annotation is prohibitively expensive and difficult to scale for complex reasoning tasks. Consequently, data synthesis is the key to breaking this deadlock.

This introduces a deeper challenge: \textbf{How to ensure synthetic financial data is both professional and reliable?} Given the strict standards of finance, simple text imitation is insufficient. We identified three core challenges:

\begin{enumerate}
    \item \textbf{Diversity \& Coverage:} Financial scenarios are vast, requiring diversity in task types (compliance, hallucination detection, numerical reasoning, table reasoning) and domains (banking, securities, insurance, accounting, macroeconomics).
    \item \textbf{Determinism of Answers:} Tasks like derivatives pricing or statement calculation require absolute numerical precision. A single error can have severe consequences. Generating reliable ``Gold Standard'' answers for synthetic instructions is critical.
    \item \textbf{Depth of Reasoning:} Many problems require multi-hop logic across multiple knowledge points. Synthetic data must move beyond shallow imitation to embody deep financial logic.
\end{enumerate}

\begin{figure}[!h]
    \centering
    \includegraphics[width=0.5\linewidth]{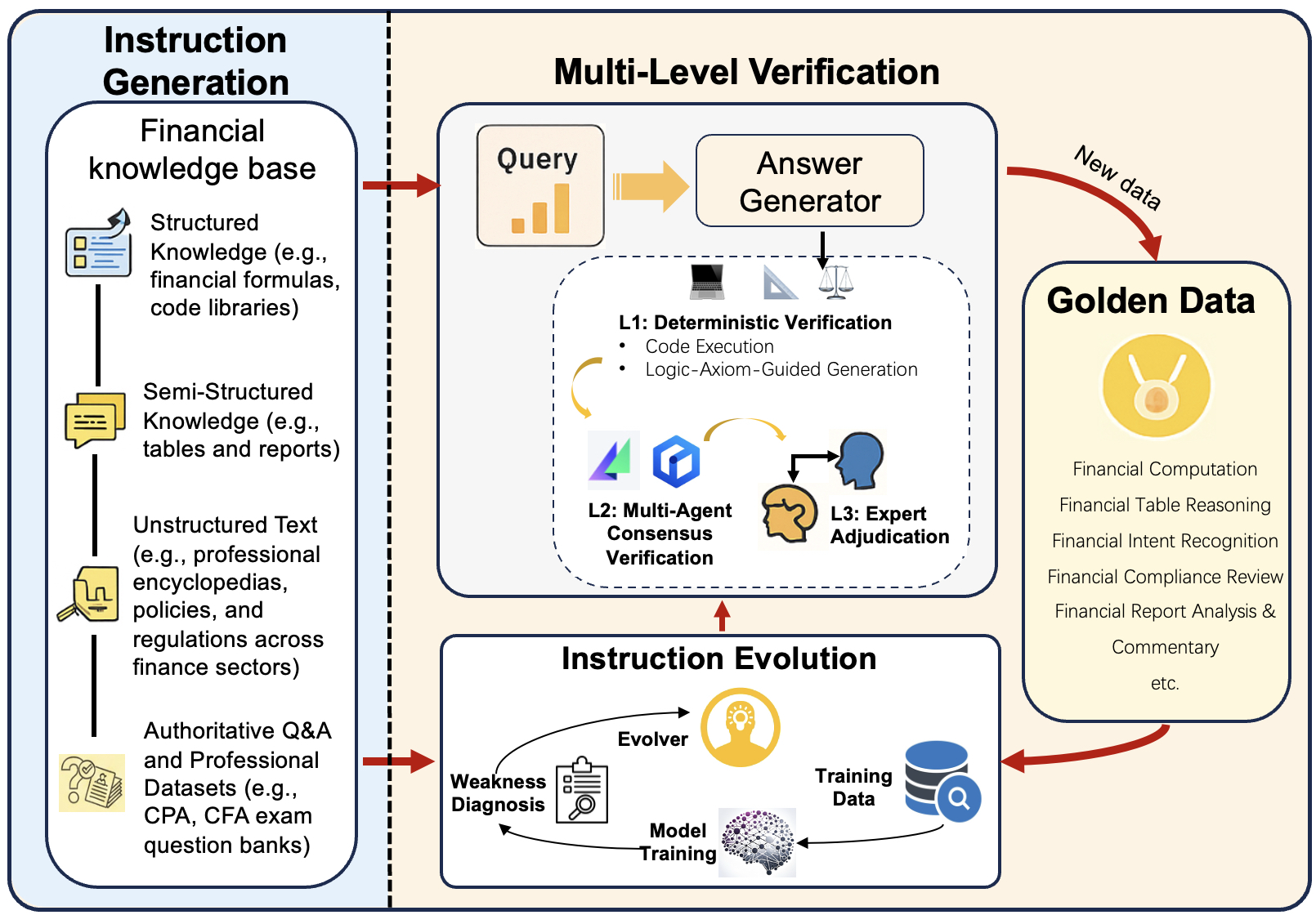}
    \caption{Overview of Data Synthesis Strategy}
    \label{fig:rl_syn}
\end{figure}

To systematically address these challenges, we propose the \textbf{Controllable Instruction Synthesis Framework for Finance (CIS-F)}, as illustrated in Figure \ref{fig:rl_syn}. This framework comprises three core modules, each specifically engineered to overcome the aforementioned data limitations.

\subsubsection{Knowledge-Driven Instruction Generation}
To solve the coverage challenge, we start at the source: the Knowledge Base. We aggregated and structured a massive volume of financial data to build a comprehensive ``Seed Knowledge Base,'' including:
\begin{itemize}
    \item \textbf{Structured:} Codebases for over 3,000 financial functions and formulas for key indicators.
    \item \textbf{Semi-structured:} Tens of thousands of financial research reports containing tables and context.
    \item \textbf{Unstructured:} Professional encyclopedias and regulations covering banking, insurance, and accounting.
    \item \textbf{Authoritative Q\&A:} Exam banks (CPA, CFA) representing high-standard domain requirements.
\end{itemize}
We employ classification models to tag raw data (e.g., ``Banking'', ``Actuarial'') and utilize \textbf{Context Injection} and \textbf{Template Guidance} to generate specific task types (e.g., ``Calculation``, ``Commenting``, ``Compliance``). This ensures the training data is not only large-scale but also highly aligned with real-world financial scenarios, preventing model capability bias.

\subsubsection{Multi-Level Cross-Verification}
To satisfy the strict accuracy requirements, we designed a verification funnel to assign a high-confidence standard answer to every synthetic instruction, abandoning the ``majority vote'' approach for a more rigorous hierarchy:

\begin{itemize}
    \item \textbf{L1: Deterministic Verification (Highest Priority):} The cornerstone of reliability.
    \begin{itemize}
        \item \textit{Code Execution:} For calculation tasks, Python code is executed to produce indisputable answers.
        \item \textit{Axiom-Driven Generation:} We embed financial formulas (e.g., $Assets = Liabilities + Equity$) into the generation logic. For instance, the system hides ``Liabilities'' based on the formula and generates a deduction problem. This ensures correctness is backed by financial theorems.
    \end{itemize}
    \item \textbf{L2: Multi-Model Voting:} When deterministic verification is unavailable, we utilize an ensemble of leading LLMs. Answers are accepted only when there is high consensus on both the result and the reasoning process.
    \item \textbf{L3: Human-in-the-Loop Adjudication:} For complex, subjective tasks (e.g., financial report commentary), domain experts are introduced to label and adjudicate, ensuring extreme quality.
\end{itemize}

\subsubsection{Instruction Evolution}
To address the reasoning depth, we introduced an \textbf{``Evolution Module.''} Instead of blindly adding data, it operates on a \textbf{``Diagnose-Evolve-Reinforce''} loop:
\begin{itemize}
    \item \textbf{Step 1: Weakness Diagnosis.} The model processes a batch of tasks. We analyze failure cases to pinpoint breaks in reasoning chains (e.g., the model succeeds in single-step ROE calculation but fails in multi-period composite calculations).
    \item \textbf{Step 2: Targeted Evolution.} The ``Evolver'' modifies existing problems by adding reasoning constraints, introducing distraction information, or transforming task formats to generate harder, more diverse variants.
\end{itemize}

\subsection{Training Strategy}

\subsubsection{SFT Training}
In the SFT stage, for samples requiring explicit reasoning, the thought process is enclosed within \texttt{<think>} and \texttt{</think>} tags. For samples not requiring explicit reasoning, the tags remain empty. The objective of SFT is to establish foundational language understanding, logical reasoning capabilities, and correct output formatting.

\subsubsection{RL Training}

\paragraph{\textbf{Reward Validation Design}}
\begin{itemize}
    \item \textbf{Reasoning RL:} Primarily utilizes \textbf{Rule-based rewards} to strictly enforce logical and computational correctness.
    \item \textbf{Agentic RL:} Uses a \textbf{Composite Reward Function} evaluating final answer correctness, tool call necessity, efficiency (number of steps), and parameter accuracy. We constructed a simulated environment where the model must clarify intent and complete information via multi-turn dialogue before invoking tools, mirroring real business scenarios.
    \item \textbf{General RL:} Given the mix of specialized financial tasks (calculation, intent recognition, commentary) and general tasks, we introduced a \textbf{Dual-Verifier System}:
    \begin{enumerate}
        \item \textbf{Rule-based Verifier:} Checks quantifiable tasks (e.g., calculation) for correctness and consistency using explicit rules and domain knowledge.
        \item \textbf{LLM-based Verifier:} Checks semantic-driven tasks (e.g., intent recognition, text inference) for semantic alignment and logical consistency.
    \end{enumerate}
\end{itemize}

\noindent\textbf{Case Study: Financial Report Commenting}\\
For tasks like financial report commenting, which demand both numerical accuracy and analytical depth, we deeply integrate both verifiers:
\begin{itemize}
    \item \textbf{Rule-based Verifier (Fact \& Format):}
    \begin{itemize}
        \item \textit{Factual Accuracy:} Extracts numerical values from the generated commentary and compares them strictly against a ``Ground Truth'' set pre-extracted from the report. Higher accuracy yields higher rewards.
        \item \textit{Formatting:} Enforces industry conventions (e.g., using ``pct'' instead of ``\%'' for YoY changes, correct list separators).
    \end{itemize}
    \item \textbf{LLM-based Verifier (Logic \& Depth):}
    \begin{itemize}
        \item \textit{Consistency (Hallucination Detection):} Semantically compares the model's analytical claims against the source text to penalize unsupported inferences or hallucinations.
        \item \textit{Structure:} Classifies paragraph themes (e.g., ``Profitability'', ``Future Outlook'') to ensure the output matches the expected professional structure.
        \item \textit{Style:} Scores the overall tone for professional conciseness and logical coherence.
    \end{itemize}
\end{itemize}

\subsubsection{Training Optimization}
To enhance generalization and efficiency, we implemented specific optimization strategies:

\begin{itemize}
    \item \textbf{Difficulty-Stratified Curriculum Learning:} We assess initial sample complexity using a \texttt{pass@10} metric. Training progresses from ``Core Knowledge'' (high pass rate) to ``Frontier Challenges'' (low pass rate), preventing gradient collapse in early stages.
    \item \textbf{Adaptive Sample Filtering:}
    \begin{itemize}
        \item \textbf{Invalid Gradient Pruning:} Following DAPO~\citep{yu2025dapo}, samples with zero reward variance (either ``saturated'' easy samples or ``impossible'' hard samples) contribute nothing to policy updates. These are dynamically pruned from the batch to purify gradient signals.
        \item \textbf{Mastery Pool Undersampling:} Samples consistently solved correctly are moved to a ``Mastery Pool.'' During batch construction, these are sampled with low probability (e.g., 0.2) solely to prevent catastrophic forgetting, while 80\% of resources focus on the ``Learning Zone'' at the model's capability boundary.
    \end{itemize}
\end{itemize}

\section{Benchmarks and Evaluation Results}

\subsection{Benchmarks Evolved}
To provide a comprehensive and objective assessment of the QianfanHuijin model, we evaluated its performance across three core dimensions: \textbf{Knowledge Capability}, \textbf{Reasoning \& Calculation Capability}, and \textbf{Agentic Capability}.

\subsubsection{Knowledge Capability Benchmarks}
\begin{itemize}
    \item \textbf{FinanceIQ~\citep{financeIQ2023}:} A comprehensive Chinese financial evaluation dataset covering 10 major categories and 36 sub-categories (e.g., CPA, CFA, Banking, Securities). It contains 7,173 multiple-choice questions assessing professional knowledge.
    \item \textbf{FLAME-Cer~\citep{guo2025flame}:} A holistic financial LLM evaluation system, designed to assess LLM professional capabilities across financial certifications: Covers 14 authoritative financial certifications (e.g., CPA, CFA, FRM) with approximately 16,000 human-verified questions. We conducted the evaluation using its \textbf{publicly available white-box test set consisting of 700 questions}.
    \item \textbf{Fin-Eva~\citep{fineva2023}:} Focuses on multi-scenario application capabilities, encompassing five dimensions: Cognitive, Knowledge, Logic, Content Generation, and Safety \& Compliance.
    \item \textbf{Fineval~\citep{guo2025fineval}:} A benchmark designed to test professional competency and safety, covering academic knowledge, industry insights, and multimodal financial tasks.
\end{itemize}

\subsubsection{Reasoning and Calculation Benchmarks}
\begin{itemize}
    \item \textbf{FinanceReasoning~\citep{tang2025financereasoning}:} An English benchmark specifically for numerical reasoning. It categorizes problems into three difficulty levels: Simple (data extraction), Medium (basic formulas), and Hard (multi-step weighted calculations and concept reasoning).
    \item \textbf{FinQA~\citep{chen2021finqa}:} A large-scale dataset focusing on numerical reasoning over hybrid data (financial texts and tables). Models must perform multi-step calculations to derive answers from corporate reports.
\end{itemize}

\subsubsection{Agentic Capability Benchmarks}
\begin{itemize}
    \item \textbf{FinIntentR-Bench (Financial Intent Recognition):} To evaluate intent understanding, we constructed a specialized benchmark. We curated samples from open sources (OpenFinData~\citep{openfindata2023}, Fin-Eva) and supplemented them with real-world business scenarios labeled by financial experts.
\end{itemize}

\subsection{Evaluation Results}

\subsubsection{Thinking Mode Performance}
We compared QianfanHuijin (Thinking Mode) against state-of-the-art models, including DeepSeek-R1 and Qwen3 
. As shown in Table~\ref{tab:thinking_results}, QianfanHuijin demonstrates significant advantages, particularly in reasoning and agentic tasks.

\textbf{Key Observations:}
\begin{enumerate}
    \item \textbf{Knowledge Dominance:} QianfanHuijin exhibits exceptional performance across all knowledge-intensive benchmarks. Notably, QianfanHuijin-70B achieves \textbf{94.43} on FLameCer and \textbf{89.59} on FinanceIQ, surpassing both DeepSeek-R1 and Qwen3-235B. This consistent superiority attributes to our extensive \textbf{Financial CPT} and high-quality \textbf{SFT} strategies.
    \item \textbf{Reasoning Breakthrough:} In the \textbf{FinQA} benchmark, QianfanHuijin-70B achieves a score of \textbf{77.1}, significantly outperforming DeepSeek-R1 (65.5). This substantial gap validates the effectiveness of our \textbf{Reasoning RL (Stage 2)}, which specifically strengthens numerical computation and multi-hop logic.
\end{enumerate}
\begin{table}[H]
    \centering
    \caption{Performance Comparison in \textbf{Thinking Mode} across Knowledge, Reasoning, and Agent Dimensions}
    \label{tab:thinking_results}
    \resizebox{\textwidth}{!}{
    \begin{tabular}{l|cccc|cc|c|cc}
        \toprule
        \multirow{2}{*}{\textbf{Model}} & \multicolumn{4}{c|}{\textbf{Knowledge Capability}} & \multicolumn{2}{c|}{\textbf{Reasoning}} & \multicolumn{1}{c}{\textbf{Agent}} \\
         & FLameCer & FinanceIQ & Fin-Eva & Fineval & FinReason & FinQA & FinIntentR & \textbf{Average} \\
        \midrule
        \multicolumn{8}{c}{\textit{70B+ Scale Models}} \\
        \midrule
        DeepSeek-R1 & 89.00 & 86.18 & 92.10 & 88.35 & 82.14 & 65.50 & 74.66 & 82.56 \\
        Qwen3-235B & 90.14 & 86.82 & 90.70 & \textbf{90.96} & 82.90 & 63.50 & 73.28 & 82.61  \\
        DeepSeek-V3.2 & 88.71 & 85.14 & 91.89 & 89.40 & 84.71 & 72.2 & 74.66 & 83.82 \\
        \textbf{QianfanHuijin-70B} & \textbf{94.43} & \textbf{89.59} & \textbf{92.53} & 86.72 & \textbf{86.72} & \textbf{77.10} & \textbf{76.31} & \textbf{86.20} \\
        \midrule
        \multicolumn{8}{c}{\textit{8B Scale Models}} \\
        \midrule
        Qwen3-8B & 78.10 & 75.51 & 87.32 & 75.07 & 75.37 & 65.60 & 74.19 & 75.88 \\
        DianJin-R1-7B & 77.20 & 76.18 & 87.92 & 79.84 & 54.14 & 56.70 & 70.95  & 71.85\\
        \textbf{QianfanHuijin-8B} & \textbf{86.14} & \textbf{83.10} & \textbf{90.10} & \textbf{81.84} & \textbf{82.60} & \textbf{68.30} & \textbf{78.24} & \textbf{81.47} \\
        \bottomrule
    \end{tabular}
    }
\end{table}

\subsubsection{Non-Thinking Mode Performance}
For latency-sensitive scenarios, we evaluated the models in Non-thinking mode, focusing primarily on knowledge retrieval and rapid QA. As shown in Table~\ref{tab:non_thinking}, QianfanHuijin maintains a leading position in domain knowledge benchmarks.

\begin{table}[H]
    \centering
    \caption{Performance Comparison in \textbf{Non-Thinking Mode} (Knowledge Focused)}
    \label{tab:non_thinking}
    \begin{tabular}{lcccc}
        \toprule
        \textbf{Model} & \textbf{FLameCer} & \textbf{FinanceIQ} & \textbf{Fineval} & \textbf{Average} \\
        \midrule
        DeepSeek-V3.2 & 86.70 & 81.97 & 88.96 & 85.88 \\
        DeepSeek-V3 & 88.71 & 82.65 & 87.31 & 87.60 \\
        Qwen3-235B (Non-Thinking) & 89.14 & 83.51 & 89.74 & 88.26 \\
        \textbf{QianfanHuijin-70B (Non-Thinking)} & \textbf{94.71} & 83.20 & \textbf{90.96} & \textbf{89.69} \\
        \midrule
        Qwen3-8B (Non-Thinking) & 76.57 & 70.64 & 75.41 & 74.21 \\
        \textbf{QianfanHuijin-8B (Non-Thinking)} & \textbf{78.43} & \textbf{81.58} & \textbf{82.28} & \textbf{80.76} \\
        \bottomrule
    \end{tabular}
\end{table}

\subsubsection{Ablation Study: Impact of Post-Training Stages}
To verify the contribution of our progressive post-training paradigm, we conducted an ablation study on the QianfanHuijin-8B model using the \textbf{FinanceReasoning} benchmark. We specifically analyzed the impact of the Reasoning RL (Stage 2) and Agentic RL (Stage 3) phases.

As shown in Table~\ref{tab:ablation_study}, the results validate our design choices:
\begin{enumerate}
    \item \textbf{Effectiveness of RL Stages:} The introduction of Reasoning and Agentic RL stages yields a substantial performance boost, elevating the score from 62.23 to 82.60 (\textbf{+20.37 points}). This indicates that the specialized RL training successfully internalized complex financial logic that SFT alone could not capture.
    \item \textbf{Synergy with Tools:} Enabling the Code Interpreter further enhances the score to \textbf{87.98}. This demonstrates that our Agentic RL stage not only improves intrinsic reasoning but also effectively aligns the model to leverage external computational tools for higher precision.
\end{enumerate}

\begin{table}[H]
    \centering
    \caption{Ablation Analysis on FinanceReasoning (Huijin-8B). We compare the baseline (SFT only) against the full pipeline and the agentic setup.}
    \label{tab:ablation_study}
    \begin{tabular}{lcl}
        \toprule
        \textbf{Configuration} & \textbf{Stages} & \textbf{Score} \\
        \midrule
        Baseline (Thinking) & w/o Stage 2 \& 3 & 62.23 \\
        QianfanHuijin-8B(Thinking) & w/ Stage 2 \& 3 (w/o tool) & 82.60$_{\textcolor{green}{\textbf{+20.37}}}$ \\
        QianfanHuijin-8B(Thinking) & w/ Stage 2 \& 3 (w/ tool) & 87.98$_{\textcolor{green}{\textbf{+25.75}}}$ \\
        \bottomrule
    \end{tabular}
\end{table}

\section{Conclusion}
In this report, we systematically presented the construction and validation of QianfanHuijin, a series of Large Language Model tailored for the financial industry. Addressing the critical challenges of general LLMs in finance---namely, domain knowledge deficits, weak reasoning chains, lack of tool synergy, and compliance risks---we proposed a complete technical stack covering data engineering, continual pre-training, post-training, and evaluation.

First, we constructed a high-quality financial knowledge corpus and implemented a Two-Stage Continual Pre-training (CPT) strategy, achieving a smooth transition from financial knowledge injection to financial capability enhancement. 

Second, we introduced a novel Progressive Post-training Paradigm (SFT $\rightarrow$ Reasoning RL $\rightarrow$ Agentic RL $\rightarrow$ General RL). This approach significantly boosted the model's capabilities in financial reasoning, complex task planning, and agentic tool collaboration. The accompanying Controllable Instruction Synthesis Framework (CIS-F) and Dual-Verifier Reward Model provided a solid foundation of high-confidence data and reinforcement signals, greatly improving the efficiency and reliability of the RL stage.

Empirical results across authoritative benchmarks demonstrate that QianfanHuijin significantly outperforms both similarly sized general models and leading industry counterparts. These results confirm the effectiveness and advancement of our methodology in depth of knowledge, breadth of reasoning, and computational precision.

Although significant progress has been made, there remains room for sustainable optimization in financial LLMs. Future work will focus on expanding financial knowledge coverage and refining model performance in niche vertical scenarios, aiming to build a financial foundation model that is trustworthy, interpretable, and deployable.

\newpage












\clearpage

\bibliographystyle{unsrt}
\bibliography{main}

\clearpage

\beginappendix

\section{Qualitative Evaluation Cases}
\label{app:cases}

To demonstrate the capabilities of \textit{QianfanHuijin} in specialized financial domains, we present three representative qualitative examples. These cases cover complex reasoning, numerical extraction from financial reports, and structured instruction following. 

\textit{Note: Since the model is pre-trained on a Chinese-centric financial corpus, the original prompts were administered in Chinese. We provide English translations for clarity.}

\subsection{Case Study 1: Investment Decision Making (Reasoning)}
This case evaluates the model's ability to apply the Capital Asset Pricing Model (CAPM) and perform chain-of-thought reasoning to make investment recommendations.

\paragraph{Prompt (Translated):}
In the following financial environment: the risk-free rate is 5\%, and the expected market return is 12\%. There are two security options: Security A has an expected return of 10\% with a Beta coefficient of 1.1, while Security B has an expected return of 17\% with a Beta coefficient of 1.2. In this scenario, which investment decision should the investor make?

\begin{modelthink}
First, I need to determine which security the investor should choose: A or B.
\begin{itemize}
    \item \textbf{Parameters:} $R_f = 5\%$, $R_m = 12\%$. Security A ($10\%$, $\beta=1.1$). Security B ($17\%$, $\beta=1.2$).
\end{itemize}

\textbf{Step 1: CAPM Analysis}
I should calculate the theoretical expected return for each security using the CAPM formula: $E(R_i) = R_f + \beta_i (R_m - R_f)$.
\begin{itemize}
    \item Market Risk Premium = $12\% - 5\% = 7\%$.
    \item \textbf{Security A:} $5\% + 1.1 \times 7\% = 12.7\%$.
    \item \textbf{Security B:} $5\% + 1.2 \times 7\% = 13.4\%$.
\end{itemize}
\textbf{Comparison:}
\begin{itemize}
    \item Security A: Given 10\% $<$ CAPM 12.7\%. It is overvalued (underperforming).
    \item Security B: Given 17\% $>$ CAPM 13.4\%. It is undervalued (provides excess return).
\end{itemize}

\textbf{Step 2: Alpha Calculation}
Let's check the Alpha ($\alpha$) for confirmation.
\begin{itemize}
    \item $\alpha_A = 10\% - 12.7\% = -2.7\%$ (Negative).
    \item $\alpha_B = 17\% - 13.4\% = +3.6\%$ (Positive).
\end{itemize}

\textbf{Step 3: Unit Risk Premium Check}
Another angle is to check the risk premium per unit of Beta.
\begin{itemize}
    \item Security A: $(10\% - 5\%) / 1.1 \approx 4.5\%$.
    \item Market: $7\% / 1 = 7\%$.
    \item Security B: $(17\% - 5\%) / 1.2 = 10\%$.
\end{itemize}
Since Security B offers a higher risk premium per unit of Beta (10\%) than the market baseline, it is the superior choice.

\textbf{Conclusion:} The investor should choose Security B.
\end{modelthink}

\begin{modelresponse}
Under these circumstances, the investor should choose \textbf{Security B}. Below is the detailed analysis:

\begin{enumerate}
    \item \textbf{CAPM Validity Check:}
    \begin{itemize}
        \item \textbf{Formula:} $E(R_i) = R_f + \beta_i (R_m - R_f)$.
        \item \textbf{Security A:} The required return is $12.7\%$. The actual expected return ($10\%$) is lower than the theoretical value, indicating insufficient compensation for risk.
        \item \textbf{Security B:} The required return is $13.4\%$. The actual expected return ($17\%$) exceeds the theoretical value, indicating it provides excess returns.
    \end{itemize}
    
    \item \textbf{Alpha Calculation (Excess Return):}
    \begin{itemize}
        \item $\text{Alpha}_A = -2.7\%$ (Negative, Underperforming).
        \item $\text{Alpha}_B = +3.6\%$ (Positive, Outperforming).
    \end{itemize}

    \item \textbf{Risk Premium Comparison:}
    Security B offers a risk premium per unit of Beta of \textbf{10\%}, which significantly outperforms the market average of 7\%.
\end{enumerate}

\textbf{Conclusion:} Security B is the superior choice due to its positive Alpha and higher risk-adjusted return.
\boxed{B}
\end{modelresponse}

\subsection{Case Study 2: Financial Statement Analysis (Numerical Extraction)}
This case tests the model's ability to extract specific numerical values from unstructured financial reports, perform percentage calculations, and provide qualitative analysis.

\paragraph{Prompt (Translated):}
Please analyze the data provided in the financial report below and answer the question.
\textit{[Context: Excerpt from China Film Co., Ltd. Q1 2023 Report]}
\\
\textbf{Question:} Calculate the contribution ratio of non-recurring profit and loss to the net profit attributable to shareholders of the listed company for Q1 2023, and analyze its impact on net profit.

\begin{modelthink}
\textbf{Step 1: Data Extraction}
I need to locate the specific figures from the provided financial report text:
\begin{itemize}
    \item Non-recurring P\&L sum: Found "Total Non-recurring Items" = \textbf{21,193,050.28 CNY}.
    \item Net Profit Attributable to Shareholders: Found in "Key Financial Data" = \textbf{181,662,559.98 CNY}.
\end{itemize}

\textbf{Step 2: Calculation}
Now, I calculate the ratio:
\[ \text{Ratio} = \frac{21,193,050.28}{181,662,559.98} \approx 0.11665... \]
Converting to percentage: approximately \textbf{11.67\%}.

\textbf{Step 3: Impact Analysis}
\begin{itemize}
    \item \textbf{Magnitude:} 11.67\% is a moderate proportion. It contributes to the profit but is not the dominant factor.
    \item \textbf{Comparison:} The "Deducted Non-net Profit" (Net profit excluding non-recurring items) is roughly 160 million.
    \item \textbf{Growth Rate Check:} The Deducted Non-net Profit grew by 92.68\% YoY, which is higher than the headline Net Profit growth (56.89\%).
\end{itemize}
\textbf{Inference:} This indicates that the company's growth is primarily driven by the recovery of its core business (cinema box office, etc.) rather than one-time gains (government subsidies, etc.).
\end{modelthink}

\begin{modelresponse}
\textbf{Non-recurring Profit \& Loss Analysis:}

\textbf{1. Calculation Steps:}
\begin{itemize}
    \item Non-recurring P\&L: \textbf{21,193,050.28 CNY} (Extracted from "Total Non-recurring Items").
    \item Net Profit Attributable to Shareholders: \textbf{181,662,559.98 CNY}.
    \item Contribution Ratio:
    \[ \frac{21,193,050.28}{181,662,559.98} \times 100\% \approx \mathbf{11.67\%} \]
\end{itemize}

\textbf{2. Impact Analysis:}
\begin{itemize}
    \item \textbf{Moderate Contribution:} Non-recurring items account for approx. 11.67\% of net profit. While beneficial, they are not the primary driver of profitability.
    \item \textbf{Core Business Strength:} The net profit excluding non-recurring items (Deducted Non-net Profit) grew by \textbf{92.68\%} YoY, significantly outpacing the headline net profit growth of 56.89\%.
\end{itemize}
\textbf{Conclusion:} The quality of earnings is high. The growth is primarily driven by the recovery of the core film business rather than one-time gains.
\end{modelresponse}

\subsection{Case Study 3: Marketing Insight (Structured Output)}
This case evaluates the model's capability in \textbf{Instruction Following} and \textbf{Customer Profiling}.

\textit{\textbf{Note:} Unlike the previous cases, this task was executed using the non-thinking mode. This demonstrates the model's ability to directly generate strictly formatted JSON for downstream agentic workflows without the latency of chain-of-thought generation.}

\paragraph{Prompt (Translated):}
Based on the input customer situation, recommend product tags for the customer. The result must only output JSON.
\\
\textbf{Constraint:} Use only the specified tags (Config: Monetary, Fixed Income...; Product: Personal Deposit, Mutual Fund...).
\\
\textbf{Customer Summary:} 
\begin{itemize}
    \item \textbf{Profile:} 38-year-old married business owner in Shanghai (High Net Worth).
    \item \textbf{Assets:} Monthly AUM of 3.37 million CNY; funds mostly in wealth management products.
    \item \textbf{Behavior:} Frequent mobile banking user; recent large-volume fund trading (risk-tolerant).
    \item \textbf{Needs:} Potential need for business liquidity and risk protection.
\end{itemize}

\begin{modelresponse_raw}
\begin{lstlisting}[style=jsonstyle]

{
    "config": [
        "Monetary Class", 
        "Fixed Income Class", 
        "Equity Class"
    ],
    "product": [
        "Personal Deposit", 
        "Public Wealth Management", 
        "Mutual Fund", 
        "Private Insurance"
    ]
}
\end{lstlisting}
\end{modelresponse_raw}


\section{Critical Review}
Explicit Statement: 
We performed rigorous n-gram decontamination to remove overlap between our training set and the test benchmarks.

\newpage

\section*{Contributors}

\textbf{\large Core Contributors}

\begin{multicols}{2}
    \begin{itemize}[noitemsep, topsep=0pt, leftmargin=*] 
        \item Shupeng Li\textsuperscript{*}
        \item Weipeng Lu
        \item Linyun Liu
        \item Chen Lin
        \item Shaofei Li
        \item Zhendong Tan
        \item Hanjun Zhong
        \item Yucheng Zeng
        \item Chenghao Zhu
        \item Mengyue Liu
        \item Daxiang Dong
        \item Jianmin Wu\textsuperscript{**}
    \end{itemize}
\end{multicols}

\vspace{1em} 

\textbf{\large Contributors}

\begin{multicols}{2}
    \begin{itemize}[noitemsep, topsep=0pt, leftmargin=*]
        \item Yunting Xiao
        \item Annan Li
        \item Danyu Liu
        \item Jingnan Zhang
        \item Licen Liu
        \item Dawei Yin\textsuperscript{**}
        \item Dou Shen\textsuperscript{**}

    \end{itemize}
\end{multicols}

\vspace{2em}

\noindent \textsuperscript{*} Project Lead \\
\textsuperscript{**} Project Sponsor

\end{document}